\documentclass[runningheads]{llncs}

\usepackage{graphicx}
\usepackage{amsmath}
\usepackage{amssymb}
\usepackage{booktabs}
\usepackage{xcolor}
\usepackage{multirow}
\usepackage[T1]{fontenc}
\usepackage{graphicx,verbatim}
\begin{document}
%
\title{\underline{Ma}mba \underline{G}uided Bo\underline{u}ndary \underline{P}rior Matters: A New Perspective for Generalized Polyp Segmentation}
%
\author{Tapas K. Dutta\inst{1} \and
Snehashis Majhi\inst{2} \and
Deepak Ranjan Nayak\inst{3} \and Debesh Jha \inst{4}}
\authorrunning{Dutta et al.}
%
\institute{University of Surrey, United Kingdom
\and
 Côte d'Azur University, France 
\and
 Malaviya National Institute of Technology Jaipur, India \and
 University of South Dakota, USA\\
\email{\color{blue}{drnayak.cse@mnit.ac.in}}\\
\url{\color{red}{https://github.com/deepak1113/SAM-MaGuP}} }



\maketitle              
\begin{abstract}

Polyp segmentation in colonoscopy images is crucial for early detection and diagnosis of colorectal cancer. However, this task remains a significant challenge due to the substantial variations in polyp shape, size, and color, as well as the high similarity between polyps and surrounding tissues, often compounded by indistinct boundaries. While existing encoder-decoder CNN and transformer-based approaches have shown promising results, they struggle with stable segmentation performance on polyps with weak or blurry boundaries. These methods exhibit limited abilities to distinguish between polyps and non-polyps and capture essential boundary cues. Moreover, their generalizability still falls short of meeting the demands of real-time clinical applications. To address these limitations, we propose \textbf{SAM-MaGuP}, a groundbreaking approach for robust polyp segmentation. By incorporating a boundary distillation module and a 1D-2D Mamba adapter within the Segment Anything Model (SAM), SAM-MaGuP excels at resolving weak boundary challenges and amplifies feature learning through enriched global contextual interactions. Extensive evaluations across five diverse datasets reveal that SAM-MaGuP outperforms state-of-the-art methods, achieving unmatched segmentation accuracy and robustness. Our key innovations—a Mamba-guided boundary prior and a 1D-2D Mamba block—set a new benchmark in the field, pushing the boundaries of polyp segmentation to new heights.


\end{abstract}

\section{Introduction}
Colorectal Cancer (CRC) is one of the leading causes of cancer-related death across the globe, ranking among the third most frequently diagnosed cancer \cite{bernal2012towards,siegel2023colorectal}.  It starts from certain growths or polyps in the inner lining of the colon and often leads to CRC if not treated timely. The incidence of CRC can be reduced significantly with early diagnosis and on-time treatment, saving many lives. Colonoscopy is the gold standard screening procedure for identifying and removing polyps. However, this detection procedure is manual, time-consuming, and highly reliant on the clinician's skill and expertise. Additionally, the miss rate of polyp detection is considerably high (6-27\%) \cite{ahn2012miss}. Therefore, it is of utmost significance to design automated polyp segmentation approaches to detect the polyp and provide clinicians with precise location and boundary information, reducing the miss rate.

There has been remarkable progress in the development of polyp segmentation methods based on deep learning in the past years, and the results shown have been awe-inspiring \cite{ronneberger2015u,zhou2018unet++,fan2020pranet}. In early efforts, U-Net and its variants, such as  UNet++ and ResUNet++ \cite{jha2019resunet++}, have been adopted as a model of choice. However, these models fall short of capturing boundary details. 
Following these works, various notable architectures such as FCN \cite{fang2019selective}, PraNet \cite{fan2020pranet},  MSNet \cite{zhao2021automatic}, CFA-Net \cite{zhou2023cross}, and MEGANet \cite{bui2024meganet} have been proposed. While such architectures exhibit their capability to capture the boundary cues and handle variability in the size of polyps, they encounter challenges in extracting global contextual information, which is crucial for polyp identification.
Later, inspired by the remarkable success of transformers, a few architectures such as PVT-Cascade \cite{rahman2023medical} and CTNet \cite{xiao2024ctnet} has recently been introduced, which allows modeling global feature dependencies and learning prominent features and have demonstrated impressive segmentation results. 

Despite the significant progress, segmenting polyps has remained a daunting challenge due to the camouflage characteristics of polyps, i.e., intrinsic resemblance among polyps and the surrounding mucosa, considerable variability in polyp's shape and size, and absence of sharp boundaries. This results in unstable segmentation performance and poor localization of polyps. In addition, the existing models have shown limited capacity to learn distinguishable features and detailed boundary cues from polyp regions. Segment Anything Model (SAM), a foundational model, has recently garnered the interest of researchers for segmenting images with exceptional generalization capabilities \cite{kirillov2023segany}. However, its performance is limited when applied to medical images because of inadequate domain-specific knowledge \cite{zhou2023can}. Additionally, it leads to high computational overhead and memory requirements \cite{wu2023medical}. To address these issues, an adapter module integrated SAM architecture has recently been proposed by leveraging Mamba~\cite{Dutta_2025_WACV}, an efficient approach known for its excellent capability to model long-range interactions with linear computational complexity \cite{zhu2024vision}. Although this approach has obtained superior generalization performance with exceptional computational benefits, it does not take into account boundary information, leading to suboptimal polyp segmentation performance. Additionally, the generalizability needs to be improved further to meet real-time clinical requirements. 

To address the persistent challenges in polyp segmentation, we introduce \textbf{SAM-MaGuP}, a novel framework that leverages boundary distillation and the Mamba adapter to enhance SAM's performance. The key building block \textbf{MaGuP} module revolutionizes general-adapter-based~\cite{adapter} fine-tuning by refining SAM’s feature representations to capture subtle boundary details often overlooked by conventional models. The module is composed of two transformative components, \textbf{1D-2D Mamba:} A unique feature fusion technique that merges channel (1D) and spatial (2D) information, empowering SAM to learn global contexts at multiple scales, unlocking richer feature representations. \textbf{Boundary Distillation Component (BDC):} A dynamic unit that sharpens the model’s ability to detect exact boundary locations, even in ambiguous or weakly defined regions. Extensive experiments across five diverse datasets demonstrate that SAM-MaGuP effectively tackles blurry boundary issues, setting a new standard by outshining current state-of-the-art (SoTA) methods. In summary, our major contributions are:
\begin{itemize} 
\item We introduce a Mamba-guided boundary prior in SAM, the first of its kind, to effectively tackle the weak-boundary challenge in polyp segmentation. 
\item We propose a boundary distillation component that empowers SAM to accurately detect polyps in weak-boundary scenarios, enhanced by a 1D-2D Mamba block that optimizes feature interactions across spatial and channel dimensions.
\item We demonstrate the outstanding performance of SAM-MaGuP on five diverse datasets, consistently surpassing existing SoTA polyp segmentation techniques. 
\end{itemize}

\section{Proposed SAM-MaGuP}
\vspace{-0.3cm}

In this section, we present the \textbf{SAM-MaGuP} framework, specifically tailored to tackle the challenging task of weak-boundary polyp segmentation. By leveraging the versatility of the SAM backbone, we elevate its capabilities to address complex segmentation tasks. We first introduce SAM, a powerful model designed for general-purpose object segmentation, before diving into the core of \textbf{SAM-MaGuP}, which enhances SAM's performance in polyp segmentation through the innovative \textbf{MaGuP Module}. This module brings unprecedented precision to weak-boundary scenarios, ensuring more effective segmentation.

\vspace{-0.2cm}
\subsection{SAM Overview and Limitations for Polyp Segmentation}

SAM comprises three key components, \textbf{Image Encoder}: Built on the ViT-H/16 architecture and pre-trained with Masked Autoencoders (MAE), this encoder uses windowed attention and global attention blocks to output down-sampled image embeddings.
\textbf{Prompt Encoder}: This module processes both sparse and dense prompts, tailored with positional encoding and learned embeddings. \textbf{Mask Decoder}: A modified transformer decoder with dynamic mask prediction, refining the interaction between image and prompt embeddings to output prediction.

While SAM excels in general object segmentation, its performance falters in weak-boundary polyp segmentation due to two major limitations: \textbf{Inferior Transfer Learning}: SAM's traditional full fine-tuning approach results in overfitting or feature degradation when trained on small or limited datasets. \textbf{Weak-Boundary Challenge}: SAM is ill-equipped to handle ambiguous, low-contrast polyp regions that blend with the background, making it difficult to generate accurate prompts or define subtle boundaries. Thus, using the original SAM for segmenting weak-boundary polyps remains a challenge.

\begin{figure}[t]
\begin{center}
\includegraphics[width=\linewidth]{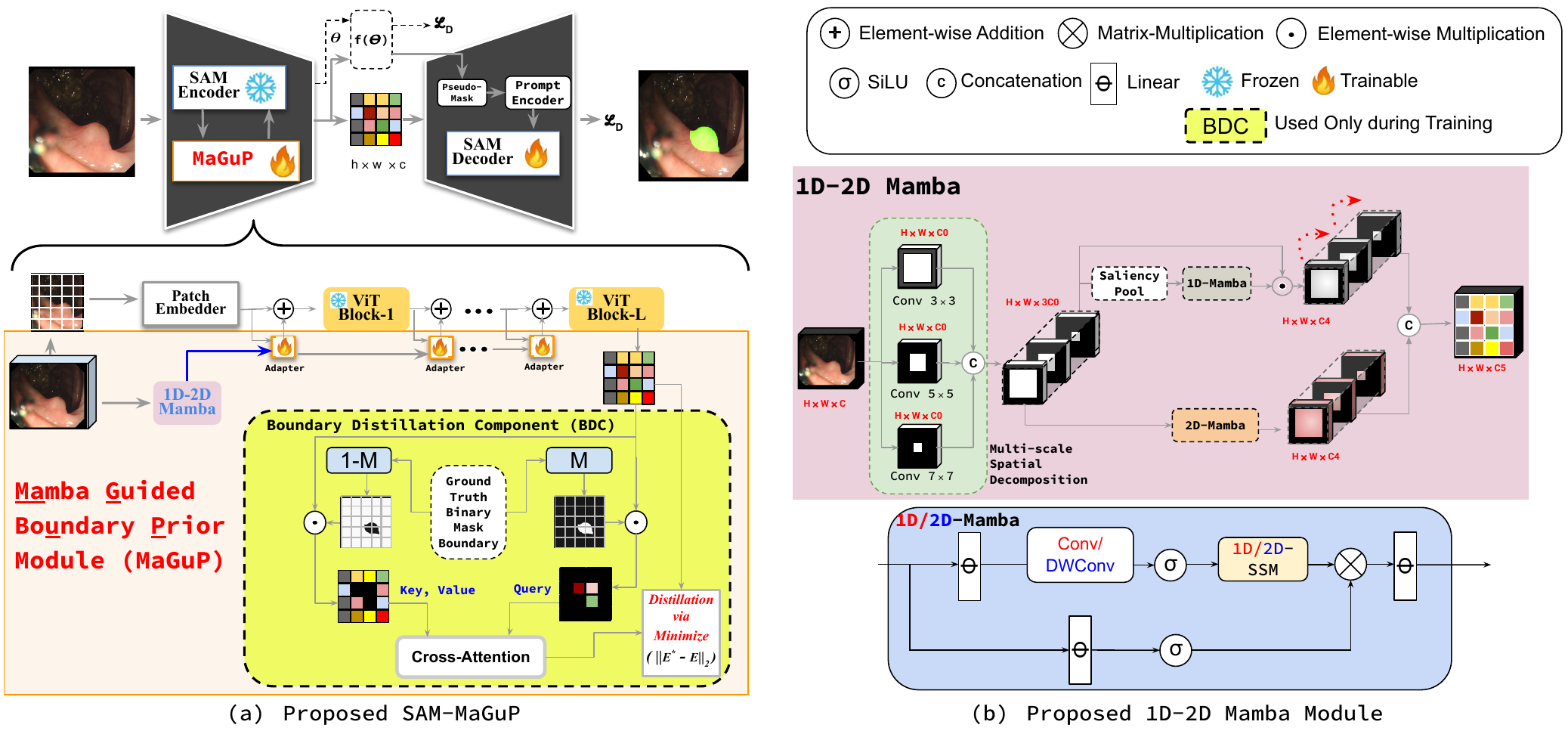}
\vspace{-0.5cm}
\caption{\textbf{Overall framework of the SAM-MaGuP} for weak-boundary polyp segmentation. It incorporated the MaGuP Adapter-based fine-tuning into the SAM backbone to enhance its representation ability for weak-boundary polyp segmentation. In MaGuP, the BDC is only used during training to refine SAM's feature learning ability pertaining to diverse polyp shapes and sizes.}
\label{fig:KDSAM}
\vspace{-0.5cm}
\end{center}
\end{figure}

\subsection{Mamba Guided Boundary Prior (MaGuP) Module}

The \textbf{MaGuP Module} introduces a groundbreaking approach to general-adapter-based fine-tuning, improving SAM's ability to tackle polyp segmentation. It acts as a plug-in general-adapter, refining SAM's feature representations to capture intricate boundary details that are often missed in traditional models. The module consists of two key components: \textbf{1D-2D Mamba}: A feature transformation that bridges the gap between channel (1D) and spatial (2D) representations, enabling SAM to better fuse channel and spatial information in global contexts for more accurate segmentation. \textbf{BDC}: A unit focused on sharpening boundary representations by transferring knowledge of precise edge localization, even in weak or ambiguous regions.

\vspace{-0.4cm}

\subsubsection{1D-2D Mamba:}

To enhance SAM's pre-trained features for polyp segmentation, we introduce the \textbf{1D-2D Mamba} {\color{red}(Fig. 1b)}. This component incorporates two powerful strategies: \textbf{Multi-scale Spatial Decomposition (MSD)}: Decomposes the input image into multiple scales through parallel convolutions with varying receptive fields, creating a multi-scale pyramid with coarse-to-fine features. \textbf{Mamba Spatial and Channel Interaction}: Captures long-range dependencies across channel and spatial dimensions by  1D and 2D Mamba layers, enhancing the ability to inject multi-scale polyp-specific cues into SAM’s encoder for improved segmentation accuracy.
\vspace{-0.2cm}

\paragraph{Multi-scale Spatial Decomposition (MSD):} An input image \(I \in \mathbb{R}^{H \times W \times C}\) is processed through parallel convolutions with varying receptive fields (\(k \in \{3, 5, 7\}\)) to capture features at multiple scales. The resulting maps \(F_k\) are padded and stacked along the channel dimension to form a multi-scale feature pyramid \(F^* \in \mathbb{R}^{H \times W \times 3C_0}\), with coarse-grained features (\(F_7\)) at the top and fine-grained features (\(F_3\)) at the bottom. This decomposition allows the model to analyze the polyp region with fine-to-coarse spatial semantics.
\vspace{-0.2cm}

\paragraph{Mamba Spatial and Channel Interaction:} Mamba efficiently captures intra-pixel interactions of salient (\(F^S\)) and contextual (\(F^C\)) maps using two parallel 1D and 2D Mamba layers \cite{zhu2024vision,liu2025vmamba}, each encoding dependencies in the multi-scale channel and spatial distributions. A gated mechanism refines feature representations, where the outputs are computed via a linear layer (\(\phi(.)\)), SiLU activation (\(\sigma\)), and matrix multiplication (\(\otimes\)). To retain fine-grained cues of channels, the original multi-scale feature maps are combined with the gated outputs using a skip connection, forming a domain-specific embedded feature map \(F^D \in \mathbb{R}^{H \times W \times C_5}\). However, the skip connection is discarded for the spatial stream to avoid information redundancy. These interaction encodings enhance the model's potential to capture critical polyp cues across scales.
\vspace{-0.4cm}

\subsubsection{Boundary Distillation Component (BDC):}
The \textbf{BDC}, as shown in \textcolor{red}{Fig. 1a}, refines SAM’s ability in to understand polyp shapes and structures. It uses the ground truth mask to clearly define polyp ($E_P$) and non-polyp ($E_{NP}$) regions, enhancing segmentation precision. By starting with the latent feature embeddings from the ViT Block-L ($E$), we compute two activation regions:

\begin{equation}
E_P = E \odot M, \quad E_{NP} = E \odot (1 - M)    
\end{equation}

where \( E \in \mathbb{R}^{h\times w \times c} \) is the feature embedding, \( M \) is the ground truth mask, and \( 1 - M \)  represents the complement of the ground truth mask (M), highlighting non-polyp regions. 

To further refine the boundary details, we employ a \textbf{cross-attention mechanism} where \( E_P \) serves as the \texttt{query}, and \( E_{NP} \) as the \texttt{key} and \texttt{value}. The resulting embedding, \( E^* \), captures the critical polyp boundaries while suppressing non-polyp regions. This refined boundary information is then integrated into SAM’s representation space using a \textbf{knowledge distillation framework}, aligning the original embedding \( E \) with the boundary-enhanced embedding \( E^* \). This is done by minimizing their representational difference:

\begin{equation}
   \mathcal{L}_{distill}= \frac{1}{h * w * c} \sum_{i=1}^{h} \sum_{j=1}^{w} \sum_{k=1}^{c} \left( E_{i,j,k} - E^*_{i,j,k} \right)^2 
\end{equation}

By minimizing this loss, the \textbf{BDC} can be discarded during inference, eliminating the need for explicit ground truth while still maintaining high boundary precision in polyp segmentation. This makes the process more efficient and accurate, even without relying on ground truth data.

\vspace{-0.3cm}
\subsection{SAM Decoder}
\vspace{-0.1cm}
The architecture of the mask decoder is adopted from~\cite{kirillov2023segany}, which leverages various prompts like bounding boxes, masks, or points for improved segmentation. However, this requires explicit prompts. To address this, \textbf{SAM-MaGuP} generates a pseudo mask via $f(\theta)$, trained with \textbf{MaGuP}, optimizing the model using $\mathcal{L}_D$ (discussed in Section~\ref{xx}). This pseudo mask subsequently serves as a prompt to refine the final segmented output during training, involving the image encoder and mask decoder.

\vspace{-0.5cm}
\subsection{Training Objective} \label{xx}
\vspace{-0.1cm}
The SAM-MaGuP model, including its components like general-adapter, MaGuP, $f(\theta)$, and mask decoder, is jointly trained with a loss \( \mathcal{L}_D = \mathcal{L}_w^{\text{Dice}} + \mathcal{L}_w^{\text{BCE}} \). For a predicted segmentation mask ($P$) and the ground truth mask ($M$), $\mathcal{L}_w^{\text{Dice}} = 1 - \frac{2 \cdot \sum (P \cdot M)}{\sum P + \sum M + \epsilon}
$, and $ \mathcal{L}_w^{\text{BCE}} = -\frac{1}{N} \sum_{i=1}^{N} \left[ M_i \cdot \log(P_i) + (1 - M_i) \cdot \log(1 - P_i) \right]$, where $N$ indicates number of pixels.

We follow a two-stage training approach:

\textbf{Stage I:} Train adapters in the image encoder with deep supervision, using up-sampled outputs \( O^{\text{Up}}_{Enc} \) and ground-truth \( M \), with loss: $\mathcal{L}_{\text{Stage-I}} = \mathcal{L}_D(M, O^{\text{Up}}_{Enc})$. \textbf{Stage II:} Train both the image encoder’s adapter and mask decoder with full supervision, using the outputs \( O_{Dec} \) and \( O^{\text{up}}_{Enc} \), with loss: $\mathcal{L}_{\text{Stage-II}} = \mathcal{L}_D(M, O_{Dec})$.

\vspace{-0.3cm}

\section{Experiments and Results}
\vspace{-0.3cm}
This section presents the quantitative and qualitative comparisons with SoTA methods, ablation study, and implementation details. 
\vspace{-0.5cm}
\subsection{Datasets, Evaluation Measures, and Implementation Details}
\textbf{Datasets:}

We conduct a series of experiments on five widely used and challenging benchmark datasets: CVC-ClinicDB~\cite{bernal2015wm}, ETIS~\cite{silva2014toward}, CVC-ColonDB~\cite{tajbakhsh2015automated}, CVC-300 (EndoScene)~\cite{vazquez2017benchmark}, and Kvasir-SEG~\cite{jha2020kvasir}. For a fair comparison, we follow the training settings of the previous study \cite{fan2020pranet}, where a mixture of images (1450 images) from Kvasir-SEG (900 images) and CVC-ClinicDB (550 images) datasets has been taken into account to train the model. For testing, the remaining 62 images from CVC-ClinicDB and 100 from Kvasir-SEG as well as 196, 380, and 60 images from ETIS, CVC-ColonDB, and CVC-300, are considered.
\\
\textbf{Evaluation Measures:}

To evaluate the performance of our proposed model as well as SoTA methods, we adopt six evaluation measures: F-measure ($F_{\beta}^{w}$) \cite{margolin2014evaluate}, S-measure ($S_{\alpha}$) \cite{fan2017structure}, E-measure ($E_{\phi}^{\text{max}}$) \cite{fan2018enhanced}, mean absolute error ($\mathcal{M}$), mean Dice (mDice), and mean IoU (mIoU),  following the previous works \cite{fan2020pranet,zhao2021automatic}.

\noindent \textbf{Implementation Details:} We implement our proposed SAM-MaGuP in PyTorch and conduct all experiments using an NVIDIA A100 GPU. To derive a fair comparison, we uniformly resize the input images to \( 352 \times 352 \) pixels. To augment the data, we adopt a multi-scale training strategy with scaling factors \{0.75, 1, 1.25\}. We train the model for 200 epochs using the Adam optimizer.  The initial learning rate and batch size are set to \( 1 \times 10^{-5} \) and 8, respectively.

\vspace{-0.5cm}
\subsection{Quantitative Comparison}
From the SoTA comparison {\color{red}(Table~\ref{tab:combined_performance})}, \textbf{SAM-MaGuP} consistently outperforms all existing methods across \textbf{seen} and \textbf{unseen} datasets. On \textbf{seen datasets}, SAM-MaGuP achieves the highest scores across all metrics. Notably, on \textbf{Kvasir-SEG}, it yields an mDice of 94.7\%, mIoU of 89.0\%, and \( S_\alpha \) of 95.1\%, significantly outperforming Polyp-Mamba and SAM-Mamba. Its \( S_\alpha \) (96.5\%) and \( E_\phi^{\text{max}} \) (98.3\%) reflect exceptional boundary localization. Similarly, on \textbf{ClinicDB}, it delivers an mDice of 95.3\% and mIoU of 91.3\%, reaffirming its robustness on challenging datasets. For \textbf{unseen datasets} {\color{red}(Table~\ref{tab:combined0})}, SAM-MaGuP demonstrates unparalleled generalization. It achieves the best scores across all metrics, effectively handling weak and ambiguous boundaries. On the challenging ETIS dataset, SAM-MaGuP sets a new benchmark with an mDice of 85.4\%, \( F_\beta^w \) of 86.2\%, and the lowest MAE of 1.0, showcasing its capability to segment complex polyp structures. These results highlight SAM-MaGuP’s robustness, adaptability, and SoTA performance. 
Further, compared to SAM-Mamba, SAM-MaGuP offers improved segmentation accuracy with minimal additional cost: 106M vs. 103M trainable parameters, identical inference-time parameters (103M), and a slight increase in GFLOPs (431 vs. 423).

In a nutshell, SAM-MaGuP not only excels on seen cases but also demonstrates exceptional generalization capabilities on unseen cases, consistently achieving top performance across all metrics. This highlights its robustness, superior boundary localization, and efficiency compared to existing SoTA methods.
\vspace{-0.6cm}

\begin{table*}[h!]
\centering
\caption{Comparison of quantitative results with SoTA approaches on \textbf{seen datasets}. `\textcolor{green}{Green}' and `\textcolor{blue}{Blue}' color fonts represent the best and second-best results.}
\label{tab:combined_performance}
\vspace{-0.3cm}
\resizebox{\textwidth}{!}{%
\begin{tabular}{|l||cccccc||cccccc|}
\hline
\multirow{2}{*}{Methods} & \multicolumn{6}{c|}{Kvasir-SEG (Seen)} & \multicolumn{6}{c|}{ClinicDB (Seen)} \\ \cline{2-13} 
 & mDice  & mIoU  & $F_{\beta}^{w}$  & $S_{\alpha}$  & $E_{\phi}^{\text{max}}$  & $\mathcal{M}$  & mDice  & mIoU  & $F_{\beta}^{w}$  & $S_{\alpha}$  & $E_{\phi}^{\text{max}}$  & $\mathcal{M}$ \\ \hline
\scriptsize U-Net {\textcolor{gray}{(MICCAI'15)}} \cite{ronneberger2015u}       & 81.8 & 74.6 & 79.4 & 85.8 & 89.3 & 5.5 & 82.3 & 75.5 & 81.1 & 88.9 & 95.4 & 1.9 \\
\scriptsize U-Net++ \textcolor{gray}{(TMI'19)} \cite{zhou2018unet++}       & 82.1 & 74.3 & 80.8 & 86.2 & 91.0 & 4.8 & 79.4 & 72.9 & 78.5 & 87.3 & 93.1 & 2.2 \\ 
\scriptsize SFA \textcolor{gray}{(MICCAI'19)} \cite{fang2019selective}        & 72.3 & 61.1 & 67.0 & 78.2 & 84.9 & 7.5 & 70.0 & 60.7 & 64.7 & 79.3 & 88.5 & 4.2 \\ 
\scriptsize PraNet \textcolor{gray}{(MICCAI'20)} \cite{fan2020pranet}         & 89.8 & 84.0 & 88.5 & 91.5 & 94.8 & 3.0 & 89.9 & 84.9 & 89.6 & 93.6 & 97.9 & 0.9 \\ 
\scriptsize MSNet \textcolor{gray}{(MICCAI'21)} \cite{zhao2021automatic}      & 90.7 & 86.2 & 89.3 & 92.2 & 94.4 & 2.8 & 92.1 & 87.9 & 91.4 & 94.1 & 97.2 & 0.8 \\ 
\scriptsize CFA-Net \textcolor{gray}{(PR'23)} \cite{zhou2023cross}        & 91.5 & 86.1 & 90.3 & 92.4 & 96.2 & \textcolor{blue}{2.3} & 93.3 & 88.3 & 92.4 & 95.0 & 98.9 & 0.7 \\
\scriptsize PVT-Cascade \textcolor{gray}{(WACV'23)} \cite{rahman2023medical} & 91.1 & 86.3 & 90.6 & 91.9 & 96.1 & 2.5 & 91.9 & 87.2 & 91.8 & 93.6 & 96.9 & 1.3 \\ 
\scriptsize CTNet \textcolor{gray}{(TCYB'24)} \cite{xiao2024ctnet}          & 91.7 & 86.3 & 91.0 & 92.8 & 95.9 & \textcolor{blue}{2.3} & 93.6 & 88.7 & 93.4 & 95.2 & 98.3 & \textcolor{blue}{0.6} \\ 
 
\scriptsize MEGANet \textcolor{gray}{(WACV'24)} \cite{bui2024meganet}       & 91.3 & 86.3 & 90.7 & 91.8 & 95.9 & 2.5 & 93.8 & 89.4 & 94.0 & 95.0 & 98.6 & \textcolor{blue}{0.6} \\

\scriptsize Polyp-Mamba \textcolor{gray}{(MICCAI'24)} \cite{xu2024polyp} & \textcolor{blue}{94.0} & \textcolor{blue}{88.1} & \textcolor{blue}{94.2} & 93.5 & \textcolor{green}{98.3} & \textcolor{green}{1.6} & \textcolor{blue}{94.9} & \textcolor{green}{90.7} & \textcolor{blue}{95.2} & \textcolor{green}{96.5} & \textcolor{green}{99.3} & \textcolor{green}{0.5} \\
\scriptsize SAM-Mamba \textcolor{gray}{(WACV'25)}  \cite{Dutta_2025_WACV} & 92.4 & 87.3 & 94.2 & \textcolor{blue}{93.6} & 96.1 & 2.5 & 94.2 & 88.7 & 94.3 & 95.5 & 98.2 & \textcolor{blue}{0.6} \\ \hline

\scriptsize \textbf{SAM-MaGuP (Ours)} & \textcolor{green}{94.7} & \textcolor{green}{89.0} & \textcolor{green}{95.1} & \textcolor{green}{94.2} & \textcolor{blue}{98.1} & \textcolor{green}{1.6} & \textcolor{green}{95.3} & \textcolor{green}{91.3} & \textcolor{green}{95.8} & \textcolor{blue}{96.3}& \textcolor{blue}{98.8} & \textcolor{green}{0.5} \\\hline
\end{tabular}%
}
\end{table*}

\begin{table*}[h!]
\centering
\caption{Comparison of quantitative results with SoTA approaches on \textbf{unseen datasets}. `\textcolor{green}{Green}' and `\textcolor{blue}{Blue}' color fonts represent the best and second-best results.}
\label{tab:combined0}
\resizebox{\textwidth}{!}{%
\begin{tabular}{|l||cccc||cccc||cccc|}
\hline
\multirow{2}{*}{Methods} & \multicolumn{4}{|c|}{CVC-300 (Unseen)} & \multicolumn{4}{c|}{ColonDB (Unseen)} & \multicolumn{4}{c|}{ETIS (Unseen)} \\ \cline{2-13} 
 & mDice & mIoU & $F_{\beta}^{w}$ & $\mathcal{M}$ & mDice  & mIoU  & $F_{\beta}^{w}$  & $\mathcal{M}$ & mDice  & mIoU  & $F_{\beta}^{w}$  & $\mathcal{M}$\\ 
 \hline
\scriptsize U-Net \textcolor{gray}{(MICCAI'15)} \cite{ronneberger2015u} & 71.0 & 62.7 & 68.4 & 2.2 & 51.2 & 44.4 & 49.8  & 6.1 & 39.8 & 33.5 & 36.6 & 3.6\\ 
\scriptsize U-Net++ \textcolor{gray}{(TMI'19)} \cite{zhou2018unet++} & 70.7 & 62.4 & 68.7 & 1.8 & 48.3 & 41.0 & 46.7 & 6.4 & 40.1 & 34.4 & 39.0 & 3.5 \\ 
\scriptsize SFA \textcolor{gray}{(MICCAI'19)} \cite{fang2019selective} & 46.7 & 32.9 & 34.1 & 6.5 & 46.9 & 34.7 & 37.9 & 9.4 & 29.7 & 21.7 & 23.1 & 10.9\\ 
\scriptsize PraNet \textcolor{gray}{(MICCAI'20)} \cite{fan2020pranet} & 87.1 & 79.7 & 84.3  & 1.0 & 70.9 & 64.0 & 69.6 & 4.5 & 62.8 & 56.7 & 60.0 & 3.1\\ 

\scriptsize MSNet \textcolor{gray}{(MICCAI'21)} \cite{zhao2021automatic} & 86.9 & 80.7 & 84.9 & 1.0 & 75.5 & 67.8 & 73.7 & 4.1 & 71.9 & 66.4 & 67.8 & 2.0 \\ 

\scriptsize CFA-Net \textcolor{gray}{(PR'23)} \cite{zhou2023cross} & 89.3 & 82.7 & \textcolor{green}{93.8} & 0.8 & 74.3 & 66.5 & 72.8 & 3.9 & 73.2 & 65.5 & 69.3 & 1.4\\ 
\scriptsize PVT-Cascade \textcolor{gray}{(WACV'23)} \cite{rahman2023medical} & 89.2 & 82.4 & 87.3  & 0.9 & 78.1 & 71.0 & 77.9  & 3.1 & 78.6 & 71.2& 75.9 & 1.3\\
\scriptsize CTNet \textcolor{gray}{(TCYB'24)} \cite{xiao2024ctnet} & 90.8 & 84.4 & 89.4 & \textcolor{blue}{0.6} & 81.3 & 73.4 & 80.1 & \textcolor{blue}{2.7} & 81.0& 73.4& 77.6 & 1.4\\

\scriptsize MEGANet \textcolor{gray}{(WACV'23)} \cite{bui2024meganet} & 89.9 & 83.4 & 88.2 & 0.7 & 79.3 & 71.4 & 77.9  & 4.0 & 73.9 & 66.5 & 70.2 & 3.7\\ 
\scriptsize Polyp-Mamba \textcolor{gray}{(MICCAI'24)} \cite{xu2024polyp} & \textcolor{blue}{92.1} & \textcolor{blue}{87.5} & 89.5 & \textcolor{green}{0.5} & 82.9 & 74.3 & 81.5 & \textcolor{blue}{2.7} & 82.5 & 74.7 & 76.6 & \textcolor{blue}{1.2}\\
\scriptsize SAM-Mamba \textcolor{gray}{(WACV'25)} \cite{Dutta_2025_WACV} & 92.0 & 86.1 &  88.8 & \textcolor{blue}{0.6} & \textcolor{blue}{85.3} & \textcolor{blue}{77.1} & \textcolor{blue}{85.6} & \textcolor{green}{1.7} & \textcolor{blue}{84.8} & \textcolor{blue}{78.2} & \textcolor{blue}{85.5} & \textcolor{green}{1.0}\\ \hline

\scriptsize \textbf{SAM-MaGuP (Ours)} & \textcolor{green}{92.7} & \textcolor{green}{88.0} & \textcolor{blue}{90.1} & \textcolor{green}{0.5} & \textcolor{green}{85.9} & \textcolor{green}{78.5} & \textcolor{green}{86.2} & \textcolor{green}{1.7} & \textcolor{green}{85.4} & \textcolor{green}{78.9} & \textcolor{green}{86.2} & \textcolor{green}{1.0}\\\hline
\end{tabular}}
\end{table*}
\vspace{-1.1cm}
\subsection{Qualitative Comparison}
\begin{figure}[t]
\begin{center}
\includegraphics[width=0.8\linewidth]{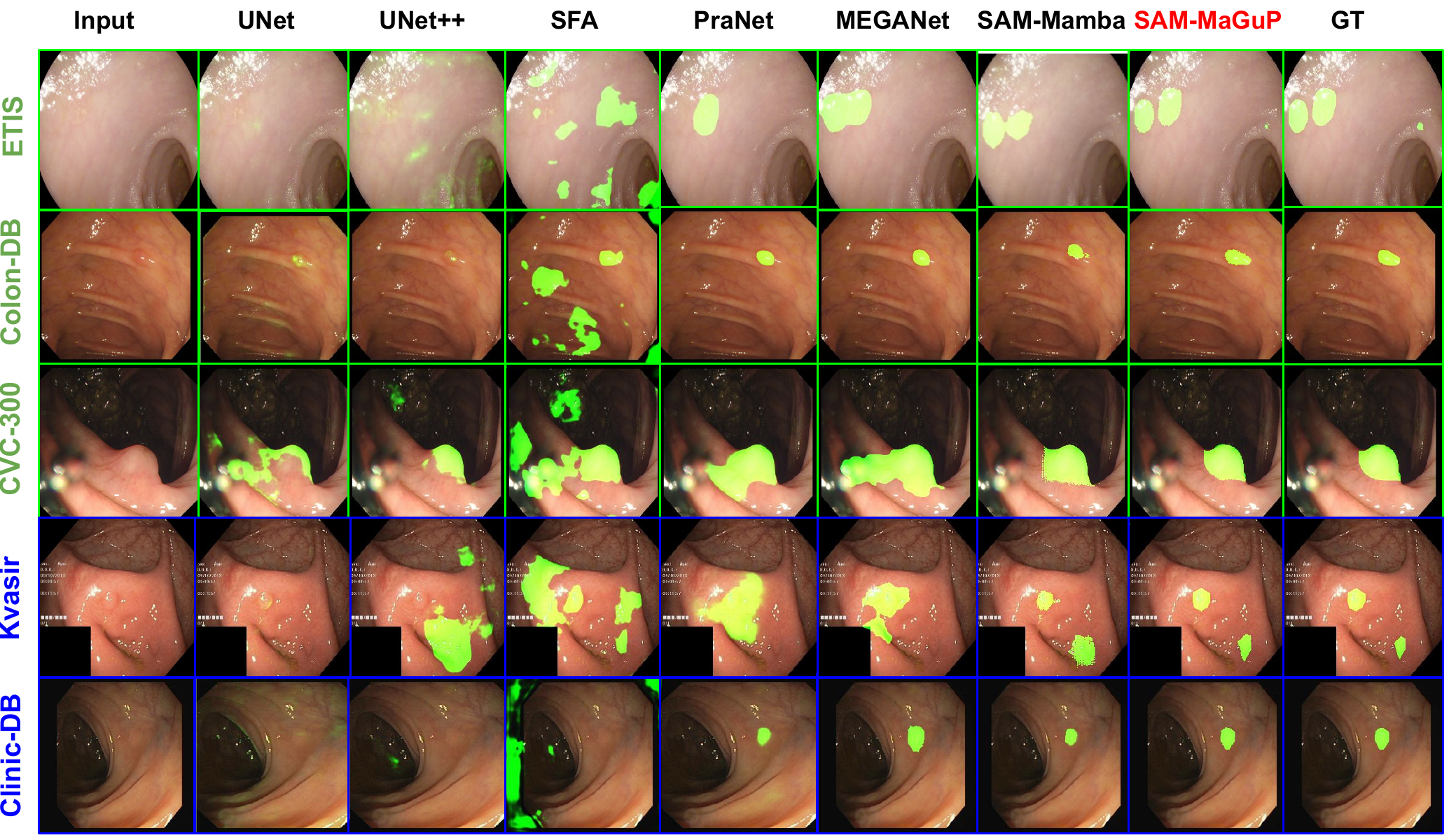}
\vspace{-0.5cm}
\caption{\small Comparison of qualitative results on both seen and unseen datasets demonstrates the SAM-MaGuP's learning ability to segment polyps with shape, size, texture variations and intricate boundaries, and showcases strong generalization.}
\label{fig:overlap}
\vspace{-1cm}
\end{center}
\end{figure}

\vspace{-0.2cm}
Qualitative evaluations {\color{red}(Fig.~\ref{fig:overlap})} demonstrate that our model excels in capturing intricate polyp boundaries across diverse datasets. Leveraging multi-scale boundary refinement, it precisely delineates polyp contours, even in low-contrast or ambiguous edge cases. On the ClinicDB dataset, SAM-MaGuP effectively distinguishes subtle boundary transitions from adjacent tissues, minimizing leakage and ensuring clear separation. In the challenging ETIS dataset, the model progressively sharpens coarse boundary cues, enhancing segmentation accuracy and reducing false positives by accurately localizing transitions between polyps and the background. These results underscore the critical role of our boundary-focused modules in delivering robust and precise polyp segmentation.

\vspace{-0.2cm}
\subsection{Ablation Study and Discussion}
\vspace{-0.2cm}

Our ablation study {\color{red}(Table \ref{table:1})} highlights the incremental contributions of each module to the overall performance of SAM-MaGuP. Across all datasets, the baseline Adapter achieves a reasonable mDice score. Subsequently, the addition of the MSD module significantly enhances multi-scale feature learning, boosting performance by an acceptable margin. Integrating the 1D-Mamba module further improves channel-wise feature discrimination, leading to an additional performance increase. Further, the 2D-Mamba module refines spatial context and local feature interactions, contributing a smaller but meaningful gain. Finally, the inclusion of the BDC sharpens boundary localization and aligns complementary features, resulting in an additional improvement. These results demonstrate the synergistic effect of each module, underscoring their critical role in achieving state-of-the-art performance.

\vspace{-0.7cm}
\begin{table}[!h]
    \caption{Results of ablation study on the components of the proposed MaGuP module.}
    \vspace{-0.1cm}
    \centering
        \resizebox{0.7\textwidth}{!}{\begin{tabular}{|c|c|c|c|c||c|c|c|c|c|}
            \hline
            \multicolumn{5}{|c|}{Configuration} & \multicolumn{2}{c|}{Seen} & \multicolumn{3}{c|}{Unseen} \\
            \cline{1-5}\cline{6-10}
            Adapter & MSD & 1D-Mamba & 2D-Mamba & BDC & Kvasir & Clinic-DB & CVC-300 & ColonDB & ETIS \\
            \hline
            \checkmark & - &- & - & - & 89.9 & 89.9 & 80.9 & 80.1 & 80.6 \\
            \checkmark & \checkmark &- & - & - & 90.9 & 91.3 & 90.3 & 80.8  & 81.2 \\
            \checkmark & \checkmark & \checkmark & - & - & 92.4 & 94.2 & 92.0 & 85.3 & 84.8 \\
           \checkmark & \checkmark &  \checkmark & \checkmark & - & 93.1 &  94.7 & 92.2 & 85.5 & 84.9\\
           \checkmark & \checkmark &  \checkmark & \checkmark & \checkmark & 94.7 & 95.3 & 92.7 & 85.9 & 85.4 \\
            \hline
        \end{tabular}}%
    
    \label{table:1}
\vspace{-0.3cm}
\end{table}

\vspace{-1cm}
\section{Conclusion}
\vspace{-0.2cm}
In summary, SAM-MaGuP represents a significant leap forward in polyp segmentation, addressing longstanding challenges related to boundary ambiguity and feature generalization. By introducing a novel BDC and a 1D-2D Mamba adapter into the SAM, SAM-MaGuP achieves precise segmentation even for polyps with weak or indistinct boundaries. Its ability to capture global contextual interactions and refine boundary cues ensures superior segmentation accuracy, setting a new standard in the field. Extensive experiments on diverse datasets confirm its robustness and clinical relevance, demonstrating that SAM-MaGuP is not only a powerful innovation in medical imaging but also a practical solution for real-world diagnostic needs. These advancements underscore the transformative potential of SAM-MaGuP in early colorectal cancer detection, offering a scalable and reliable framework for next-generation polyp segmentation.

\bibliographystyle{splncs04}

\end{document}